# Bangla Fake News Detection Based On Multichannel Combined CNN-LSTM

Md. Zahin Hossain George
Dept. of CSE
Daffodil International University
Dhaka, Bangladesh
zahin15-9710@diu.edu.bd

Naimul Hossain
Dept. of CSE
Daffodil International University
Dhaka, Bangladesh
naimul15-10036@diu.edu.bd

Md. Rafiuzzaman Bhuiyan
Dept. of CSE
Daffodil International University
Dhaka, Bangladesh
rafiuzzaman15-9655@diu.edu.bd

Abu Kaisar Mohammad Masum
Dept. of CSE
Daffodil International University
Dhaka, Bangladesh
abu.cse@diu.edu.bd

Sheikh Abujar
Dept. of CSE
Independent University
Dhaka, Bangladesh
abujar@iub.edu.bd

*Abstract*—There have recently been many cases of unverified or misleading information circulating quickly over bogus web networks and news portals. This false news creates big damage to society and misleads people. For Example, in 2019 there was a rumor that the Padma Bridge of Bangladesh needed 100,000 human heads for sacrifice. This rumor turns into a deadly position and this misleading information takes the lives of innocent people. There is a lot of work in English but a few works in Bangla. In this study, we are going to identify the fake news from the unconsidered news source to provide the newsreader with natural news or real news. The paper is based on the combination of convolutional neural network (CNN) and long short term memory (LSTM) where CNN is used for deep feature extraction and LSTM is used for detection using the extracted feature. The first thing we did to deploy this piece of work was data collection. We compiled a data set from websites and attempted to deploy it using the methodology of deep learning which contains about 50k of news. With the proposed model of Multichannel combined CNN-LSTM architecture, our model gained an accuracy of 75.05% which is a good sign for detecting fake news in Bangla.

*Index Terms*—Fake news detection, LSTM, CNN

## I. INTRODUCTION

Online content has been playing a significant role in recent years. Online platform for news is also valuable for people. So a significant amount of online newspapers are publishing day by day. To violate the peaceful world, some people are misleading the situation by spreading fake news on an online platform. It hampers peoples day to day life. An increased political diversion was the reason for the popularity of social media (eg Facebook), easier access to online advertising revenue. Some time political parties spread fake news to manipulate people to gain personal and party interest. Traditional organizations send us the real news but fake organizations send fake news from different fake platforms and media. Since to cross-check reference people are often unable to spend enough time and be sure about the credibility of news, automated detection of fake news is indispensable. Therefore, the research community is receiving great attention from it.

There are many instances where consequences had severe by instigating religious or ethnic groups against innocent victims from cleverly designed fake news. It is now very easy to spread fake news with the help of available internet and fake media in a populated country like Bangladesh. In October 2019, in Bangladesh, there was news flashing that the great bridge of Padma River needs human heads and child blood and it is growing so rapidly that many innocent people lost their lives in craze from the fake news. That's why we can use NLP and machine learning to detect fake news. By machine learning, we will learn machine that is real news or which is fake news. And one of the well-known fields in the community is natural language processing by the help of NLP we can process and manipulate human language to detect fake news. In English, there are lots of research works. But it is a big challenge for us to detect fake news in Bangla because it is very difficult to detect fake news in Bangla because of its structure. There are 342 million people around the world who speak in Bangla. So fake news can manipulate a lot of the population. So we expect that our work will do a significant change in detecting Bangla fake news. In the rest of our research paper, we will describe our data set and the methods that have been used to develop the accuracy of trials.

In this piece of work, our model multichannel combined CNN LSTM performs outstandingly in case of detecting fake news especially Bangla fake news. This model can be also used for the problems within this domain. Our data set also adds value in case of solving problems like this specifically in Bangla text classification.

## II. LITERATURE REVIEW

Acknowledging the effect of fake news many researchers are trying to detect fake news in different ways. Most of the fake news is published on online or social media. They are trying different methods to detect fake news. Usually, there are various types of fake news. One research team has proposed that there are three types of fake news. SeriousFabrications(TypeA), LargeScaleHoaxes(TypeB), HumorousFakes (TypeC) [6].

Previous works to detect fake news are mostly used the SVM method or linguistic feature. A model introduced by author Zobaer to use SVM, NLP technique with 50K article data set With an accuracy of 91% in Bangla fake news detection [1]. Another model introduced by author Aditi that uses machine learning, Random forest classifier algorithm with 500 articles of data set holding accuracy of 86% in Bangla fake news Detection [2].

Author Vivek introduced a model that is a novel computational method focused on text analysis to automatically detect fake news. They use the "Kaggle Fake News" data set provided by the SBP-BRIMS organizers with an accurate 345 data set. They evaluate the text with 5 points character count, Authentic, Clout, Tone, Analytic. Contribution (1) Development of a new public data set with relevant new articles (2) Development of text-processing computer-based learning for automated recognition of Fake News with 87% accuracy [3].

Author Aswini has suggested using neural network architecture to accurately predict the position between a specific arrangement of headlines and the body of an article. The focus of their system is to identify fake news through position detection. Automatically detecting the connection between two parts of the text is called Position Detection. Data set used: Fake News Challenge (FNC-1) Data. They achieve 94.31% accuracy with Tf-Idf vector word representations combined with pre-processed engineering features as convolutional inputs and to predict targets position they used dense neural network architecture [4].

Author Veronica suggests a model that detects false news based on linguistic features that gain accuracy of up to 78%. They build their data set, FakeNewsAMT data set, Celebrity data set. They've measured Human performance with Fivefold cross-validation with precision, recall, accuracy, and also with SVM as a classifier [5].

Author Junaed introduced their neural network-based models. They have used three Liar4, False, or Real News datasets 6, Combined Corpus. They found that among the conventional machine learning models, Naive Bayes, with n-gram (TF-IDF bigram) features, performed the best with 94% in the combined corpus. Lexical and emotional features, SVM, and LR models have performed better than other conventional machine learning models. In the neural CNN model, the best and, finally, the proposed Conv-HAN hybrid model exhibits high performance on all three datasets that undoubtedly attract interest for potential experimentation with a larger data set [6].

Author, Sherry has suggested using the Deep learning algorithm RNN(GRU) method with LIAR data set with an accuracy of 0.217 FI score [7].

Author Supanya has proposed Machine learning Naïve bays which data sets combine with fake social media news articles from Twitter. In which Neural Network & SVM method gets 99.90% accuracy [8].

Author Hadeer has suggested to use of Linear-SVM as a classifier and also TF-IDF as a function extraction strategy. With 92% accuracy, they have created their real news data from real hosts and fake news data from the Kaggle website [9].

A model proposed by author Shafayat can identify false news-based newspaper heading. Using the Gaussian Naive Bayes method, they were able to generate a new Bengali language data set and accomplish their goal of reaching the objective. Other algorithms were tried, however, the Gaussian Naive Algorithm fared well in their model. The attribute was chosen using an Extra Tree Classifier and a text feature based on TF-IDF. They achieved 87% in their model using Gaussian Naive Bayes, which is better than any other method they tested [10].

A hybrid technique proposed by author Arnab combining Word2Vec with TF-IDF to extract features from text documents. They have created their data set from various news portals like Prothomalo and Ittefaq for real news and Motikontho for satire news. They were able to determine if a Bangla text content was satire or not with a 96 % accuracy rate using their suggested feature extraction method and conventional CNN architecture [11].

Author Gulzar has proposed two machine learning techniques in their study of research. To detect Bangla false news, classifiers such as Support Vector Machine (SVM) and Multinomial Naive Bayes (MNB) were used. Around 2500 articles were gathered for their data set from ProthomAlo, BalerKontho, Motikontho and ProthomAlu news sources. Term Frequency Inverse Document Frequency Vectorizer and Count Vectorizer were used to extract features. Their proposed method detects false news based on the polarity of the relevant post. SVM with linear kernel has a 96.64%, whereas MNB has a 93.32%, according to their results [12].

We found from the literature survey that rule-based strategies worked well in false news identification by machine learning techniques. The latest techniques in deep learning can be used to achieve improved accuracy and better prediction. Driven by this, we have presented our problem set with a hybrid CNN-LSTM architecture. CNN implemented first than the LSTM layer used. This architecture has improved performance and better f1 score as well. We have used deep learning technique, Multichannel combined CNN-LSTM architecture which can execute feature engineering on its own. Without being expressly informed, it will examine the data for features that correlate and combine them to facilitate faster learning. The deep learning architecture is adaptable, which means it may be used to solve new problems in the future.

III. METHODOLOGY

To train and evaluate our model we have used Tensor flow 2.2.0. We need some pre-processing objects for our data set before training. We would then introduce our suggested model. Figure-1 shows the methodology we've used.

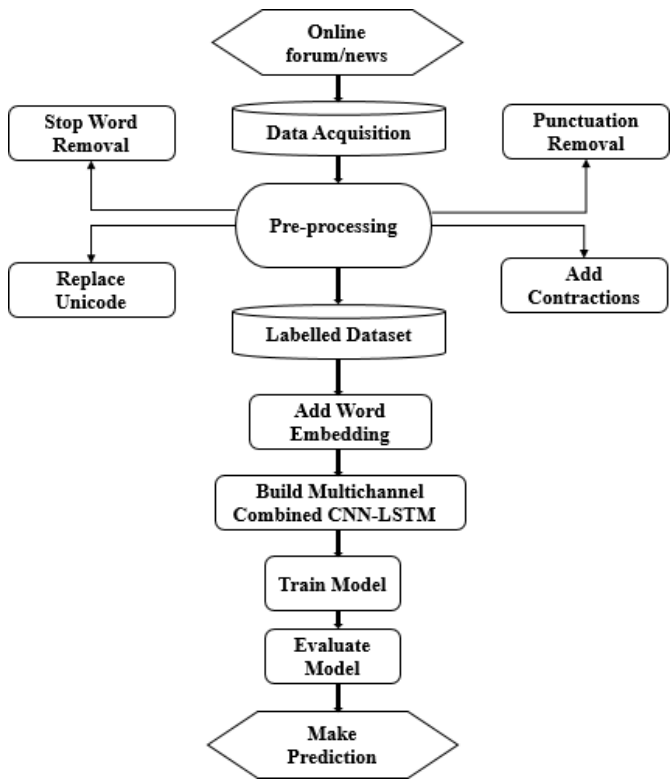

Fig. 1. Workflow of Fake-news detection.

## A. Data Collection

The method of collecting and evaluating the information on specified variables in a defined framework that helps one to answer specific questions and measure results is called data collection. It is a very important section of a paper where all can know about the origin of the data that have been used in the research. There are all about 50k of data set which we all have collected from 22 most popular and reputed news portals of Bangladesh [10]. The data set contains miss leading content, clickbait, parody news, which we have collected from jaachai.com [11] and bdfactcheck.com [12] Such two platforms offer a clear and detailed description of a still released fake news on other pages. Clickbait is one type of news that attracts newsreaders with misleading headlines, which redirect them to another website to gain their visitor number. We found that all the local newspapers do that trick. We manually collect that news from the sites. We have also collected the meta-data with the headline and also the domain of the news site, category, and publication time. Which creates all-most a number of 50k of data set.

## B. Data Pre-processing

Before working with a data set it is very important to reshape a data set to its perfect length to get the best result from it. From the data set with a length of 50k, we have reshaped the data length to get a better performance result from our suggested model. Before representing the data using a machine learning model, certain refinements such as stop-word removal, tokenization, lower case, sentence segmentation, and punctuation removal need to be subjected to the data. This will assist us to lessen the scale of real information by doing away with the inappropriate facts inside the information. We got 242 categories of news from the data set. From them, we have categorized 12 types of news from the data set. Miscellaneous news, Entertainment, Lifestyle, National, International, Politics news, Sports news. Crime, Education, Technology, Finance, and Editorial news To build our data set additional capable, we've got enclosed the supply data as a meta-data for every news. Besides that, the headline article link is also included in the meta-data. After checking the relationship between the headline and the article, tags like "Related" and "Unrelated" are given. By these processes, we have managed to collect almost 8.5k data from the data set & we labeled real news as 1 and fake news as 0 for better understanding [1]. Completion of the process gives us a better result of our data set.

## C. Feature Engineering

Texts are represented as a word vector. There is numerous study for this representations. Word embedding is a powerful way to represent words that can learn from textual data. it is a vector representation of words and can give us a powerful semantic relationship between the words. But as far we are working with Bengali so we need our word embedding. Sometimes it is a good approach to working on pre-trained embedding. So in our case "bn w2v model" was used for our purpose. Before training, we need to finalize our vocabulary size. For this, from the pre-processed text we count the vocabulary.

## D. Model

There are several types of deep learning models out there. Each of them is used for a variety of purposes. Just like ConvNet is used to handle image data and video processing stuff. In our problem set Multichannel Combined CNN-LSTM is used here. Figure-2 shows our model architecture.

*1) Embedding Layer:* Embedding layer, which is the primary layer of the model. Each sentence is known to be a sequence of $t_1, t_2, ..., t_N$ word tokens, where $N$ is the vector length of the token. According to news expression figures, approximately 95 percent of phrases are shorter than 30 sentences. We thus restrict the sum of $N$ to 30 empirically. Any sentence(news) longer than thirty tokens is truncated to 30 exploitation null cushioning and any sentence(news) lower than 30 is cushiony to 30. Any term is mapped to a word vector with a $d$-dimension. This is the matrix $T \in \Box^{N \times d}$ of the output layer.

*2) CNN Layer:* A sliding window of width w in each CNN layer where M filters are added over a matrix of the previous embedding layer. Assume that $F \in \Box^{w \times d}$ denotes a filter matrix and bias is represented by $b$. Assuming that the token vectors $t_i, t_{i+1}, ..., t_{i+j}$ (if $k > N$, $t_k = 0$) are denotes by

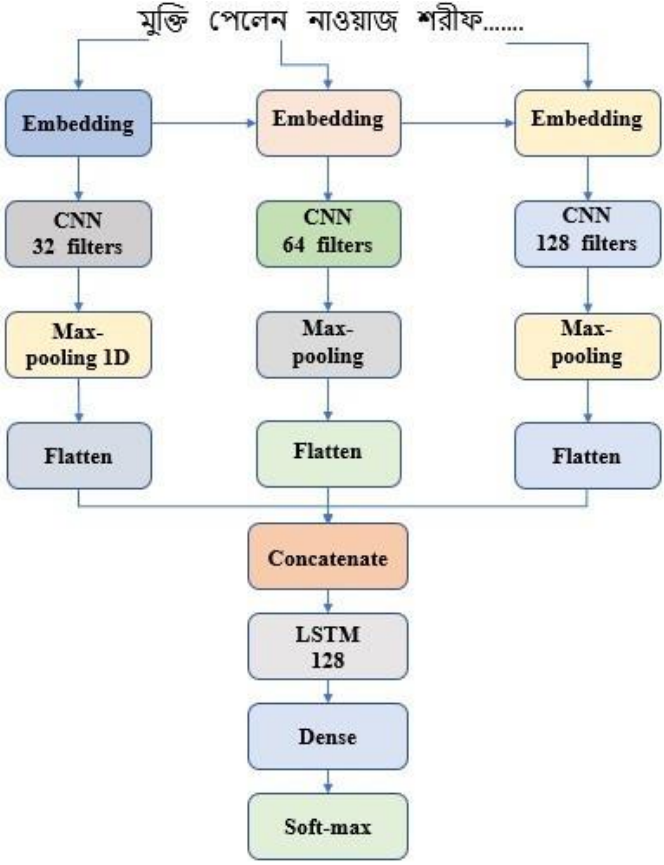

Fig. 2. Proposed model

$T_i : i + j$, each filter results will be $f \in \Box^d$ in the generation of the I variable of f by:

$$f_i = T_{(i:i+w-1)} \otimes F + b \quad (1)$$

Where the operation of convolution denotes by $\otimes$. A nonlinear activation function is introduced until f is processed to the next sheet. Here, for quicker estimation, we use the ReLU function (Nair and Hinton, 2010). The $w$ gram function can be derived by combining filters with window width $w$. We can extract active local $n - gram$ characteristics on different scales by adding several converting filters to this substrate. We apply zero padding to token vectors prior to convolution to keep the output sizes of different filters similar.

*3) Max-Pooling Layer:* Throughout this sheet, as the most popular element, the maximum value from various filters is taken. Because $W$ gram features can be retrieved by the CNN layer with window width $w$, the maximum output values from the CNN layer are considered to be the most prominent information in the targeted sentence (news). We choose max rather than mean pooling because the most defining attribute of a word (news) is the salient function.

*4) LSTM Layer:* For the analysis of sequential data, the recurrent neural network (RNN) architecture is ideal. Due to the gradient vanishing problem, however, a simple RNN is typically difficult to learn. LSTM implements a gating framework to fix this issue that enables clear memory upgrades and deliveries. LSTM uses the following formula to measure hidden state ht:

- input gate, $i_t = \sigma(W^{ix} x_t + W^{ih} h_{t_1})$
- output gate, $o_t = \sigma(W^{ox} x_t + W^{oh} h_{t_1})$
- forget gate, $f_t = \sigma(W^{fx} x_t + W^{fh} h_{t_1})$
- process input or input node, $g_t = \tanh(W^{gx} x_t + W^{gh} h_{t_1})$
- memory state, $h_t = \tanh(s_t) \otimes o_t$
- hidden node, $s_t = s_{t-1} \otimes f_t + g_t \otimes i_t$

Where $i_t$ is for input vector, $g_t$ stands for the vector of the cell state; layer parameters denote by W, U, and b ; vectors of the gate represented by $f_t$, $i_t$, and $o_t$ ; and function of the sigmoid stands for . Note that $\otimes$, actually denotes the product of Hadamard.

*5) Hidden Layer:* This is a layer that is completely linked. It multiplies outcomes with a weight matrix from the previous layer and applies a bias vector. Also added is the ReLU activation function. Finally, the result vectors are inserted into the output layer.

*6) Output Layer:* The final classification result is produced by this sheet. Using soft max as an activation feature, it is a connected network. This layer's value is a vector determined by:

$$P(y = j|x) = \frac{e^{x^T w_j}}{\sum_{k=1}^{k} e^{e_{i,j}}} \quad (2)$$

where input vector denoted by x, weight vector i denoted by w and K is the number of classes. So that, we find the final classification result $\hat{y} = argmax_j P(y = j|x)$

## IV. EXPERIMENT & RESULT ANALYSIS

We also predefined some of the parameters in our study. This is also known as hyper-parameter architecture. For the proper work of our model, we must have to declare these parameters. For the better work of our model, we first have to train them. Here table 1 shows our models hyper-parameter:

TABLE I
HYPERPARAMETERS

| Batch size | 124 |
|---|---|
| Epoch | 75 |
| learning rate | 0.0001 |
| CNN filter | 32, 64,128 |
| LSTM's size | 128 |

We train our model by Tensorflow 2.0.1. In our model, we set hyperparameters first. Then we divided our data set. The percentage of our data for training is 80 and the rest of 20% are used for testing. Our model was later compiled into an adam optimizer with a lower learning rate and added binary cross-entropy as the loss parameter. Then a callback from Earlystopping launched to track our lack of validation. This is accompanied by the .fit() process to start underway with preparation. It will stop practicing until our validity failure

TABLE II
CLASSIFICATION REPORT

| NewsType | Precision | Recall | f1-score |
|---|---|---|---|
| Real | 0.71 | 0.63 | 0.67 |
| Fake | 0.81 | 0.67 | 0.74 |
| Average. | 0.76 | 0.65 | 0.70 |

doesn't change using callbacks.

We see from the above table that our average precision of 0.76, recall of 0.65 is very good with a 0.70 f1 ranking. Since the language of Bangla is so varied and nuanced, it is quite tough to get more precision into this form of assignment.

## V. CONCLUSION & FUTURE WORK

Due to the potential impact, fake reviews can have on consumer behavior and deception detection, purchasing decisions in online reviews & fake news has an important role in business, national security, law enforcement, politics over recent years. Researchers use machine learning with a large data set to increase learning and thus get the best results by using a word embedding to extract features or signs that distinguish syntactic-semantic relationships between words. It is very important to verify which is fake news and which is real news because that impacts society so much & we have done the job with our proposed model. With the accuracy of our suggested model, we can identify fake news behind real news. The benefit of the model gives a huge impact on the area of fake journalism. If we find another scope to develop the model, it will be a huge success in the future. In this paper, we have used a machine learning algorithm to address fake news detection. On three different data sets, we present an overall performance analysis of the various approaches. We were able to structure our model successfully.

On which precision is all about 0.76. Now, this model is active in identifying fake news and actual news. Our 8.5k data set length model requires 7k data to be labeled as authentic and 1.5k data to be labeled as fake [1]. After testing the data set we get an excellent result with our proposed model. Because this is early research of Bangla's fake news detection our future work will increase the work's accuracy from a larger data set.